\documentclass[sigconf]{acmart}
\usepackage[T1]{fontenc}
\usepackage[utf8]{inputenc}
\usepackage{xcolor}
\usepackage{listings}
\usepackage{caption}

\usepackage[ruled,vlined]{algorithm2e}
\usepackage{float}

\lstdefinestyle{mypython}{
  language=Python,
  basicstyle=\ttfamily\small,
  keywordstyle=\bfseries,
  commentstyle=\itshape,
  showstringspaces=false,
  tabsize=2,
  frame=single,
  breaklines=true,
  columns=fullflexible,
  captionpos=b
}
\AtBeginDocument{%
  }

\copyrightyear{2026}
\acmYear{2026}
\setcopyright{cc}
\setcctype{by}
\acmConference[WSDM Companion '26]{The Nineteenth ACM International Conference on Web Search and Data Mining}{February 22--26, 2026}{Boise, ID, USA}
\acmBooktitle{The Nineteenth ACM International Conference on Web Search and Data Mining (WSDM Companion '26), February 22--26, 2026, Boise, ID, USA}
\acmPrice{}
\acmDOI{10.1145/3779211.3793173}
\acmISBN{979-8-4007-2358-2/2026/02}

\begin{document}

\title{SPARK: Search Personalization via Agent-Driven Retrieval and Knowledge-sharing}

\author{Gaurab Chhetri}
\affiliation{%
  \institution{Texas State University}
  \city{San Marcos, Texas}
  \country{USA}}
\email{gaurab@txstate.edu}
\orcid{0009-0000-0124-4814}

\author{Subasish Das}
\affiliation{%
  \institution{Texas State University}
  \city{San Marcos, Texas}
  \country{USA}}
\email{subasish@txstate.edu}
\orcid{0000-0002-1671-2753}

\author{Tausif Islam Chowdhury}
\affiliation{%
  \institution{Texas State University}
  \city{San Marcos, Texas}
  \country{USA}}
\email{sgp98@txstate.edu}
\orcid{0009-0008-2385-8719}

\renewcommand{\shortauthors}{Chhetri et al.}

\begin{abstract}
Personalized search demands the ability to model users’ evolving, multi-dimensional information needs; a challenge for systems constrained by static profiles or monolithic retrieval pipelines. We present SPARK (Search Personalization via Agent-Driven Retrieval and Knowledge-sharing), a framework in which coordinated persona-based large language model (LLM) agents deliver task-specific retrieval and emergent personalization. SPARK formalizes a persona space defined by role, expertise, task context, and domain, and introduces a Persona Coordinator that dynamically interprets incoming queries to activate the most relevant specialized agents. Each agent executes an independent retrieval-augmented generation process, supported by dedicated long- and short-term memory stores and context-aware reasoning modules. Inter-agent collaboration is facilitated through structured communication protocols, including shared memory repositories, iterative debate, and relay-style knowledge transfer. Drawing on principles from cognitive architectures, multi-agent coordination theory, and information retrieval, SPARK models how emergent personalization properties arise from distributed agent behaviors governed by minimal coordination rules. The framework yields testable predictions regarding coordination efficiency, personalization quality, and cognitive load distribution, while incorporating adaptive learning mechanisms for continuous persona refinement. By integrating fine-grained agent specialization with cooperative retrieval, SPARK provides insights for next-generation search systems capable of capturing the complexity, fluidity, and context sensitivity of human information-seeking behavior.
\end{abstract}




\received{27 November 2025}

\maketitle

\section {Introduction}
       
Personalized search systems have long struggled with rigidity in how they model users. In traditional approaches, a user’s interests are represented as a static profile (e.g., a vector of topic preferences) or at best a short recent session history \cite{teevan2005personalizing, dou2007large}. This static or one-dimensional view fails to capture the fluid and multifaceted nature of real information needs. Users often exhibit evolving intents and may undertake “atypical” searches that diverge from their usual profile \cite{eickhoff2013personalizing}, yet static-profile pipelines cannot easily accommodate such shifts. Moreover, conventional search architectures are largely monolithic: a single ranking algorithm or pipeline serves all queries, with limited adaptation per user or context. This one-size-fits-all paradigm makes it hard to address scenario-specific requirements, such as shifts in task or role. In sum, current personalization techniques are often too coarse-grained and inflexible, struggling with context changes and complex user objectives \cite{ontanon2021personalization_paradox,dharap1998context_based_profile}.

\begin{figure}

            \centering
            \includegraphics[width=1\linewidth]{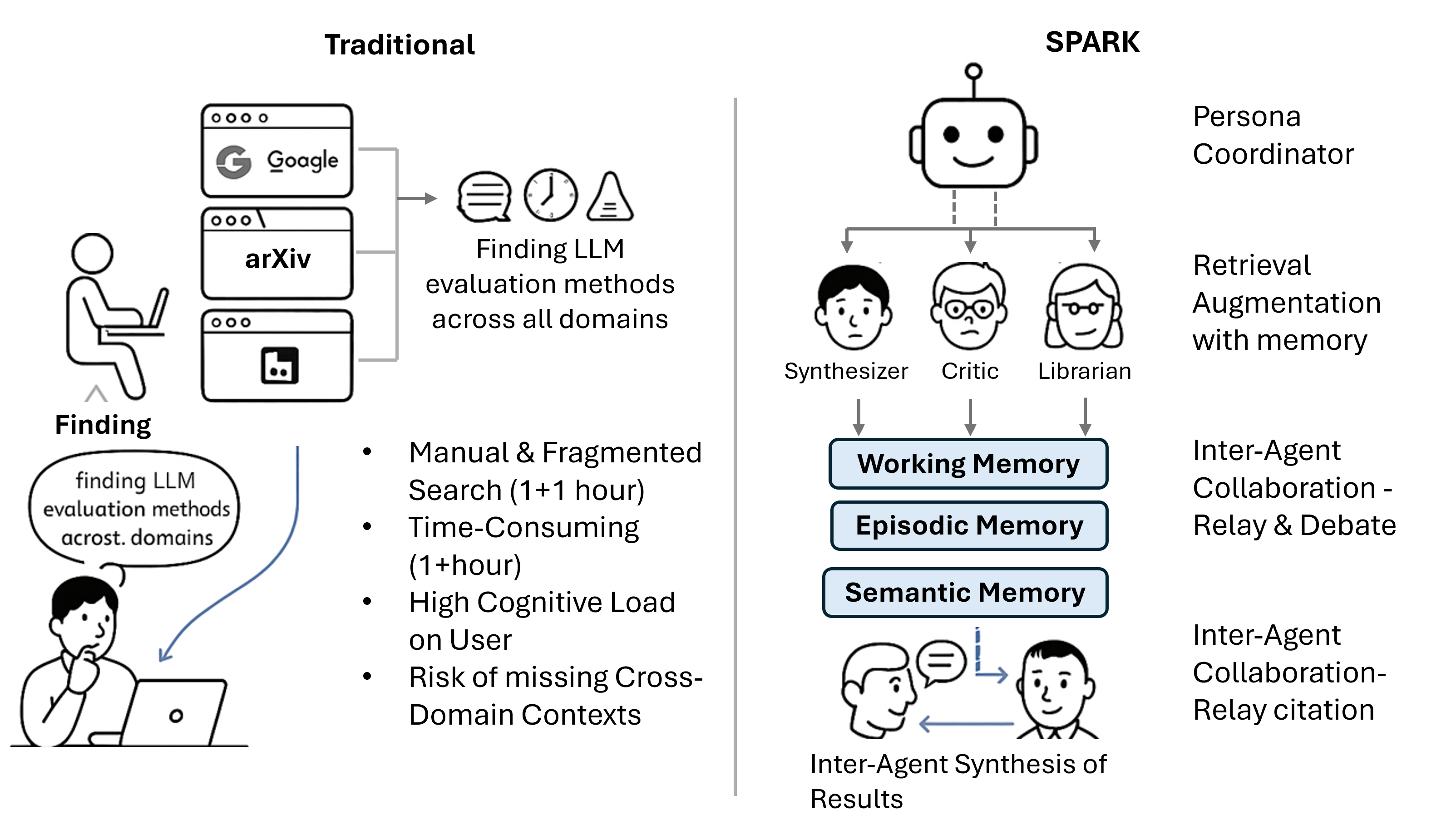}
                \caption{Traditional manual search is slow, fragmented, and cognitively demanding, whereas SPARK’s automated multi-agent approach for finding LLM evaluation methods across domains leverages specialized persona agents, layered memory, and inter-agent collaboration to produce faster, more comprehensive, and evidence-backed results.}
        \label{fig:placeholder}
\end{figure}

Recent advances in large language models (LLMs) motivate a rethinking of personalization toward dynamic, context-aware systems \citep{das_artificial_2023}. LLM-based agents can perform complex reasoning and are capable of following instructions or roles, suggesting an opportunity to decompose the search process into specialized sub-tasks \cite{bassani2022neural_personalized_query_expansion}. Instead of a single omniscient ranker, one can envision multiple coordinated agents—each with a distinct persona or expertise—working together to retrieve and synthesize information. A multi-agent approach allows personalization to become contextual and adaptive: depending on the current query and situation, different expert agents can be invoked. For example, one agent might focus on recalling a user’s long-term domain preferences while another scrutinizes results for a specific task at hand. Such agents can communicate and adjust in real time, enabling retrieval strategies that are tailored to the user’s immediate context and latent needs \cite{steichen2019adaptive_visualization}. This stands in contrast to static pipelines, offering a path to more fluid personalization. The use of LLM agents further permits advanced interaction patterns—agents can debate conflicting evidence or relay findings in stages—providing richer mechanisms to handle complex queries. In essence, dynamic multi-agent architectures promise a leap in personalization, moving beyond fixed profiles toward emergent personalization behavior that adjusts as the user’s context and goals evolve \cite{teevan2005personalizing, oliaee_automating_2025}.

In this work, we present SPARK (Search Personalization via Agent-Driven Retrieval and Knowledge-sharing), a framework that realizes this vision. SPARK introduces a formal persona model and multi-agent architecture to capture the multidimensional nature of users’ information needs. First, we define a persona formalism that characterizes agents along facets of role (e.g., “Critic” vs. “Synthesizer”), expertise (subject matter focus), task context (type of user goal), and domain, allowing fine-grained specialization of search behavior. A central Persona Coordinator analyzes each incoming query in context and activates the most relevant persona agents accordingly. Second, we develop structured coordination protocols for these agents to interact: agents may operate as independent specialists whose results are later fused, engage in a relay chain where each agent builds on the previous one’s output, or enter a constrained debate where agents critique each other’s answers under the oversight of a judge agent. These minimal protocols enable cooperation or adversarial refinement of results in a controllable manner. Third, SPARK’s design draws inspiration from cognitive architectures, incorporating separate memory stores analogous to human working and long-term memory. Each agent maintains a short-term working memory and episodic memory of recent interactions, while a shared long-term semantic memory stores durable user preferences and knowledge – a structure that mirrors classic cognitive models separating transient buffers from stable knowledge \cite{baddeley2000episodic, anderson2004integrated}. This cognitive-inspired memory design can help the system distinguish the user’s momentary context from persistent interests, supporting continuity across sessions without sacrificing adaptability. Finally, we outline an evaluation framework for SPARK, including a set of testable predictions and experimental methodologies. We propose specific hypotheses (e.g., about when multi-agent debate improves answer quality) and describe how to evaluate personalization quality, coordination efficiency, and cognitive load using both offline log simulations and user studies. Through this combination of persona-driven agents, coordinated retrieval, and cognitive memory modeling, SPARK aims to provide a next-generation personalization approach that captures the complexity, fluidity, and context-sensitivity of human information-seeking behavior.

\section {Related Works}
\subsection{Search Personalization}
Personalization in web search has often relied on static or slowly evolving user profiles, where long-term user interests are encoded from historical behavior—e.g., clicks or browsing patterns—and treated as fixed preference vectors \cite{teevan2005personalizing, radlinski2011personalizing}. Although such profiles improve relevance for recurring queries, they struggle to adapt when a user’s intent shifts unexpectedly—such as during "atypical" sessions \cite{eickhoff2013personalizing}. To address this, researchers have explored session-based personalization, which adapts to immediate query contexts, and hybrid approaches that blend long- and short-term modeling, including attention-based integration \cite{guo2018attentive,yu2019adaptive,purificato2024survey}. Fresh developments in temporal profiling leverage LLMs to fuse short- and long-term user preferences in real time \cite{sabouri2025temporal}. Nonetheless, most deployed systems implement personalization as a simple reranking mechanism using partial signals, rather than fundamentally altering retrieval based on user context \cite{dou2007large}.

\subsection{Retrieval-Augmented Multi-Agent Systems}

Retrieval-augmented generation (RAG) combines traditional information retrieval with LLM generation, where the LLM synthesizes responses grounded in retrieved external documents, thereby reducing hallucination and improving factual accuracy \cite{lewis2020retrieval, shuster2021retrieval}. While RAG has proven transformative for open-domain QA and conversational search, most implementations remain general-purpose rather than tailored to individual users. Personalization within RAG is still emerging, focusing on techniques such as injecting user-specific context into retrieval or generation, e.g., through personalized query expansion, though these approaches are not yet widely adopted. Multi-agent systems offer complementary benefits by structuring interactions among specialized agents to improve reasoning and retrieval. Debate-based architectures foster adversarial exchanges, often moderated by a judge agent, which has been shown to enhance answer consistency and correctness \cite{du2023improving, liang2023encouraging}. Relay models enable sequential refinement of queries and results, while cooperative models leverage agents with distinct domain expertise—such as planner, executor, and verifier roles—to produce more robust outputs \cite{liang2023encouraging}. Despite their promise, these paradigms are largely unexplored in personalized search and user-specific agent coordination could better address complex and evolving information needs.

\subsection{Motivation and Scope}

Personalized search often treats users as static vectors or as short session histories. That makes it hard to capture task shifts, evolving intents, and role-dependent information needs. Large language model agents let us decompose search into coordinated, persona-specialized reasoning steps that adapt during a session. SPARK proposes a principled way to do that: a persona space, a coordinator that routes queries to specialized agents, memory that separates short-term working context from long-term user knowledge, and simple coordination rules that cause useful personalization to emerge without brittle, hand-wired pipelines. We draw on three strands of prior work: retrieval-augmented generation for grounded answers, multi-agent collaboration and debate for stronger reasoning, and cognitive architectures for memory and control separation.

\subsection{SPARK’s Contributions}

SPARK’s design is grounded in cognitive models that distinguish between working memory (short-term focus), episodic memory (recent interactions), and semantic memory (long-term knowledge), drawing on theories such as Baddeley’s multi-component model \cite{baddeley2000episodic} and cognitive architectures like ACT-R (Anderson et al., 2004). While some IR systems implicitly adapt long- and short-term memory by separating session context from persistent user profiles \cite{dou2007large}, SPARK explicitly implements these memory modules and uses a Persona Coordinator to orchestrate their use—mimicking human cognitive control to deliver targeted personalization. This cognitive foundation integrates seamlessly with SPARK’s broader contributions, which unify long- and short-term user modeling, retrieval augmentation, multi-agent reasoning, and cognitive-inspired memory design into a single framework. Prior work has advanced each of these domains in isolation but has not combined them to enable dynamic, context-aware personalization. Current RAG-based personalization remains limited, multi-agent coordination is rarely coupled with deep user modeling, and cognitive-memory structures in IR are seldom operationalized at an architectural level. SPARK addresses these gaps by introducing a persona-based multi-agent system with layered memory and structured coordination patterns (relay, debate, and fusion), enabling adaptive, context-sensitive search experiences that evolve with the user’s needs.

\section{Formal Problem}

\paragraph{Define a Persona.} Let a user issue a session $s = \{(q_t, c_t)\}_{t=1..T}$ of queries $q_t$ with context $c_t$ that includes prior queries, clicks, dwell, and optional side signals. We define a persona as a 4-tuple

$$
\pi = (r, e, t, d)
$$

with role $r$ (for example, synthesizer, critic, librarian), expertise $e$ (for example, academic IR, programming, health), task context $t$ (for example, exploratory learning vs. decision making), and domain $d$ (for example, transportation safety, LLM systems). The persona space $\mathcal{P}$ is the Cartesian product of these facets with embeddings $\psi(\pi) \in \mathbb{R}^k$.

At time $t$, a Persona Coordinator computes a stochastic routing distribution over a subset of agents $A_t \subset \mathcal{A} \subseteq \mathcal{P}$:

$$
w_t(\pi) = \text{softmax}\big(W\, \phi(q_t, c_t) \cdot \psi(\pi)\big)
$$

where $\phi$ encodes query and session context. The top-k agents under $w_t$ are activated.

\paragraph{Retrieval target.} Each active agent $\pi$ returns a ranked list $R^\pi_t$ and a rationale $E^\pi_t$. An arbiter fuses $\{R^\pi_t\}$ into a final ranking $R_t$ and synthesizes the answer using the rationales. We use late fusion with calibrated confidences for stability. Reciprocal Rank Fusion is a strong default when score distributions are not well calibrated.

\section{System Architecture and Framework}

SPARK is designed as a modular, persona-driven multi-agent retrieval and reasoning system inspired by principles from cognitive architectures, multi-agent coordination theory, and information retrieval. The framework operationalizes claims about coordination efficiency, personalization quality, cognitive load distribution, and diversity, while maintaining a high degree of modularity for experimentation.

\vspace{-10pt}

\subsection{Architectural Overview}

At a high level, SPARK consists of five interdependent components: the Persona Coordinator, a set of specialized persona agents, a tripartite memory subsystem, a unified tool and data connector layer, and an Arbiter that integrates and verifies outputs. The architecture is modular by design, enabling controlled ablations and comparative studies across coordination strategies.

\noindent \textbf{Persona Coordinator:} The coordinator receives the current query $q_t$ and session context $c_t$ and determines both which agents to activate and which collaboration protocol to employ. Protocols include \textit{independent execution} (parallel retrieval without communication), \textit{relay-style reasoning} (sequential handoff between agents), and \textit{constrained debate} (iterative evidence vetting). The routing decision is guided by a learned policy, potentially implemented as a contextual bandit, that balances exploration and exploitation.

\noindent \textbf{Persona Agents:} Each persona agent $\pi_i$ is a large language model instance configured with a distinct role, knowledge scope, and reasoning style. The agent executes a retrieval-augmented generation (RAG) loop that integrates both external corpus retrieval and internal memory recall, producing structured notes and a ranked result list. By instantiating multiple agents with complementary perspectives, SPARK increases the diversity of reasoning paths and mitigates mode collapse.

\noindent \textbf{Memory Subsystem:} SPARK separates agent memory into \textit{working} ($M_s$), \textit{episodic} ($M_e$), and \textit{semantic} ($M_{\text{sem}}$) stores. Working memory holds the transient state relevant to the current step, episodic memory captures prior interaction traces for task continuity, and semantic memory encodes long-term user preferences, domain ontologies, and entity graphs. This separation is motivated by cognitive science findings that specialized memory buffers improve both efficiency and fidelity.

\noindent \textbf{Tool and Data Connectors:} Agents access external knowledge through standardized connectors, compliant with the Model Context Protocol, that wrap web search application programming interface (APIs), domain-specific datasets, and computational tools. This enables agents to retrieve heterogeneous content without hard-coded integrations, and to extend capabilities dynamically.

\noindent \textbf{Arbiter:} The Arbiter fuses outputs from multiple agents using a combination of rank aggregation and learned scoring. In addition to producing the final ranked list, the Arbiter ensures that all synthesized responses are grounded in verifiable evidence, with citations preserved. Fusion strategies can be swapped for experimentation, e.g., Reciprocal Rank Fusion vs. supervised learning-to-rank.

\subsection{Agent Execution Algorithm}

The core persona agent loop is illustrated in Algorithm~\ref{alg:persona_agent_step}. The agent integrates both external retrieval $D_{\text{web}}$ and memory-based retrieval $D_{\text{mem}}$, plans a tool usage strategy, reasons over retrieved evidence, ranks results, and updates its memory.

\begin{algorithm}[t]
\caption{Persona Agent Step with External and Internal Retrieval}
\label{alg:persona_agent_step}
\DontPrintSemicolon
\SetKwInOut{Input}{Input}
\SetKwInOut{Output}{Output}

\Input{$q$ (query), $session\_ctx$ (session context), $\pi$ (persona)}
\Output{$R$ (ranked results), $E$ (evidence)}

$M_s, M_e, M_{\text{sem}} \leftarrow \mathrm{ReadMemory}(q, session\_ctx)$\;
$D_{\text{web}} \leftarrow \mathrm{WebRetrieve}(q, \pi, session\_ctx)$\;
$D_{\text{mem}} \leftarrow \mathrm{RetrieveFrom}(M_e, M_{\text{sem}}, q)$\;
$P \leftarrow \mathrm{Plan}(q, \pi, session\_ctx)$
$E \leftarrow \mathrm{Reason}(P, D_{\text{web}} \cup D_{\text{mem}}, M_s)$\;
$R \leftarrow \mathrm{Rank}(D_{\text{web}}, D_{\text{mem}}, \pi)$\;
$\mathrm{WriteMemory}(\mathrm{UpdateFrom}(E))$\;
\Return{$R, E$}\;
\end{algorithm}

\subsection{Coordinator Algorithm}

The Persona Coordinator determines routing and collaboration mode. Algorithm~\ref{alg:coordinator} presents an instantiation where the protocol choice is influenced by predicted difficulty and novelty.

\begin{algorithm}[t]
\caption{Persona Coordinator with Protocol Selection}
\label{alg:coordinator}
\DontPrintSemicolon
\SetKwInOut{Input}{Input}
\SetKwInOut{Output}{Output}

\Input{$q_t$ (query), $c_t$ (context), $B$ (budget)}
\Output{$answer$, $R_{\text{fused}}$}

$scores \leftarrow \mathrm{PredictGate}(q_t, c_t)$
$A \leftarrow \mathrm{TopKPersonas}(scores, k\_from\_budget(B))$\;
$proto \leftarrow \mathrm{SelectProtocol}(q_t, c_t, scores)$\;

\If{$proto = \text{debate}$ \textbf{and} $\mathrm{GateConfidence}(q_t, c_t) < \theta$}{
    $proto \leftarrow \text{independent}$\;
}

\If{$proto = \text{independent}$}{
    $results \leftarrow \mathrm{ParallelMap}(\mathrm{PersonaAgentStep}, (q_t, c_t, \pi) \ \forall \ \pi \in A)$\;
}
\ElseIf{$proto = \text{relay}$}{
    $results \leftarrow \mathrm{RelayChain}(q_t, c_t, A)$\;
}
\Else{
    $results \leftarrow \mathrm{BoundedDebate}(q_t, c_t, A, r_{\max})$\;
}

$R_{\text{fused}} \leftarrow \mathrm{ReciprocalRankFusion}(results.rankings)$\;
$answer \leftarrow \mathrm{Synthesize}(q_t, results.notes, R_{\text{fused}})$\;

\Return{$answer, R_{\text{fused}}$}\;
\end{algorithm}

This architecture enables empirical tests of SPARK’s hypotheses:

\begin{itemize}

\item \textbf{Coordination efficiency (H1)} is measured by comparing utility per token across independent and debate protocols under varying task complexity.
\item \textbf{Personalization quality (H2)} is tested by tracking improvements in session utility when using contextual-bandit routing versus static routing.
\item \textbf{Cognitive load distribution (H3)} is evaluated through latency and context-length metrics when working memory is separated from long-term stores.
\item \textbf{Diversity and serendipity (H4)} is assessed using ERR-IA@k and subtopic coverage under diversity-optimized routing.
\end{itemize}

The combination of Algorithms~\ref{alg:persona_agent_step} and \ref{alg:coordinator} with the architectural components provides a concrete, falsifiable instantiation of SPARK’s framework, suitable for both offline evaluation and controlled user studies.

\subsection{Coordination Protocols}

Following guidance from \citet{smith2021multiagent}, SPARK employs three primary coordination protocols to produce multiple persona-specialized agents, each tailored to different query complexities and retrieval goals. These protocols are inspired by prior work in multi-agent information retrieval \citep{li2023camel, du2023improving, liang2023encouraging} and multi-step reasoning systems \citep{nakano2021webgpt, wang2023plan}.

\paragraph{Independent Specialists:}  
In the \textit{independent specialists} mode, the top-$k$ most relevant persona agents are activated in parallel based on an initial relevance score assigned by the Persona Coordinator. Each agent performs retrieval and synthesis independently, using its domain-specific strategies and personalized user models. The retrieved lists are subsequently fused into a consolidated ranking and final answer using RRF \citep{cormack2009rrf} or learned fusion models. This approach minimizes coordination overhead, reduces latency, and is well-suited for straightforward queries where parallel retrieval yields sufficient coverage. Similar parallel specialist models have been successfully applied in federated search \citep{arguello2011federated}, demonstrating robust performance across heterogeneous retrieval sources.

\paragraph{Relay:}  
The \textit{relay} protocol establishes a sequential chain of agents, denoted $\pi_1 \rightarrow \pi_2 \rightarrow \dots \rightarrow \pi_m$, where each agent processes the intermediate output (notes, retrieved evidence, reformulated queries) of its predecessor. This enables multi-stage reasoning: for example, an initial agent may interpret and expand the query, a second may target specialized corpora, and a third may synthesize the results into a coherent answer. Relay is particularly effective for complex, compositional information needs where stepwise refinement is critical, echoing findings from iterative query reformulation research \citep{clarke2008iterative, narasimhan2016learning}. While more computationally expensive than parallel retrieval, relay pipelines can significantly improve precision on under-specified or multi-faceted queries.

\paragraph{Constrained Debate:}  
In the \textit{constrained debate} protocol, two or more agents engage in a bounded, adversarial exchange of critiques over a fixed number of rounds. Each agent defends its own retrieved evidence while challenging the validity of others' results. A judge agent evaluates the presented evidence and reasoning, selecting the most credible answer set. This design is motivated by multi-agent debate frameworks \citep{du2023improving, liang2023encouraging} that have been shown to enhance factual accuracy and reasoning robustness in LLMs. Debate is particularly beneficial for ambiguous or high-stakes queries where reasoning correctness is paramount; however, it incurs additional latency and computational cost.

\paragraph{Trade-offs and Adaptive Selection:}  
Each protocol offers distinct benefits and drawbacks. Independent specialists maximize efficiency but may miss opportunities for cross-agent refinement. Relay pipelines improve interpretive depth but are slower. Debate protocols often yield the most rigorously vetted answers but at the highest cost. To balance these trade-offs, SPARK employs a learned gating mechanism—implemented as a contextual bandit \citep{li2010contextual}—to adaptively choose the optimal protocol based on predicted query difficulty, ambiguity, and user profile characteristics. This ensures that collaborative benefits are realized only when needed, preserving responsiveness on simpler queries.

\subsection{Learning and Adaptation}

SPARK continuously learns and adapts at three key levels: routing strategies, memory management, and persona refinement. This design ensures that the system not only optimizes current performance but also improves over time through principled online and offline learning methods.

\paragraph{Adaptive Routing with Contextual Bandits:}  
The Persona Coordinator frames the selection of both personas and coordination protocols as a contextual decision-making problem \citep{li2010contextual, agrawal2013thompson}. Given the feature vector $\phi(q_t, c_t)$—capturing the query representation $q_t$ and contextual state $c_t$ such as user history, session metadata, and task indicators—the coordinator selects an action $a_t$ (e.g., a subset of persona agents and a protocol) and observes a delayed reward signal. This reward is computed as a composite of implicit and explicit feedback signals, including click-through rate, dwell time, satisfaction ratings, and downstream task success. Algorithms such as LinUCB \citep{li2010contextual} and Thompson Sampling \citep{russo2018tutorial} are employed to balance exploration (trying less-used agents/protocols) and exploitation (leveraging historically effective choices), with additional constraints to bound cognitive load and computational latency. This adaptive routing mechanism has been shown in large-scale personalization systems \citep{chapelle2011bandits} to effectively match strategies to varying user needs.

\paragraph{Memory Update Strategies:}  
Each persona maintains a \textit{dual memory architecture} inspired by cognitive theories of working, episodic, and semantic memory \citep{baddeley2000episodic, anderson2004integrated}. Episodic memory stores recent interaction tuples—comprising queries, retrieved results, user interactions, agent notes, and inferred task states—providing context for ongoing sessions. Semantic memory encodes long-term, stable user preferences and expertise profiles, updated only when the system has high confidence in the persistence of a preference. For instance, repeated and consistent selection of Python-related code snippets across sessions leads to a stable semantic memory entry “prefers Python code examples.” This separation mirrors cognitive principles: keeping working/episodic memory lightweight and adaptable, while allowing semantic memory to accumulate structured, domain-specific knowledge over time \citep{dou2007large}.

\paragraph{Persona Refinement via Off-Policy Evaluation:}  
Each persona agent’s configuration—such as retrieval model parameters, reranking features, or prompt engineering strategies—is continuously refined. To avoid destabilizing live performance with untested changes, SPARK employs \textit{off-policy learning} \citep{swaminathan2015counterfactual}. Logged interaction data is used to simulate alternative decision policies, applying counterfactual estimators like inverse propensity scoring (IPS) and doubly robust estimation \citep{dudik2011doubly} to predict performance without affecting online metrics. For example, an academic-focused librarian persona might be evaluated on a new citation-prioritization heuristic offline; only if the evaluation shows significant improvement in relevance and trustworthiness is the change deployed online. This safe, data-driven refinement loop ensures robust, incremental improvement of each specialist agent.

\subsection{Result Fusion and Answer Synthesis}

Once the selected persona agents return their retrieval outputs, SPARK must integrate heterogeneous evidence streams into a unified, high-quality response. This process involves robust rank fusion, calibrated confidence scoring, diversification for intent coverage, and final natural language synthesis.

\paragraph{Fusion Methods and Confidence Calibration:}  
The fusion stage is coordinated by a central arbiter that merges the per-agent ranked lists into a single aggregated ordering. By default, SPARK employs \textit{Reciprocal Rank Fusion} (RRF) \citep{cormack2009rrf}, a rank aggregation method known for its stability under heterogeneous scoring functions. RRF computes the reciprocal of each document's rank position (with a tunable offset parameter) across lists and sums them to produce a fused score, ensuring that highly ranked documents from any agent contribute strongly to the final order. Since agents may operate on distinct scoring scales, the arbiter performs \textit{confidence calibration} \citep{guo2017calibration}, mapping raw scores to well-calibrated probability estimates before fusion. Confidence-weighted tie-breaking ensures that no single specialist's output dominates unfairly, while preserving the benefits of multi-agent diversity. Although RRF provides a strong, domain-agnostic baseline, the system can optionally employ a \textit{learned fusion model} \citep{karpukhin2020dense, nogueira2019multi} when sufficient labeled data is available, enabling adaptive weighting of agents based on query type and context.

\paragraph{Answer Synthesis:}  
After fusion, a \textit{synthesizer persona} constructs the final answer. This agent receives the fused top-$k$ documents or passages as structured evidence and generates a coherent, natural-language response grounded in the retrieved content. To ensure transparency, the synthesizer cites source documents inline and flags uncertain claims with hedging markers. The synthesis process incorporates techniques from retrieval-augmented generation (RAG) \citep{lewis2020retrieval} and factuality-aware prompting \citep{shuster2021retrieval} to reduce hallucinations and preserve semantic fidelity to the retrieved evidence.

\paragraph{Intent Coverage and Diversity Optimization:}  
Personalization can lead to over-narrow retrieval when user profiles dominate ranking preferences. To counteract this, SPARK optimizes for \textit{intent coverage} during fusion using the Expected Reciprocal Rank – Intent Aware (ERR-IA) metric \citep{chapelle2011intentaware}. ERR-IA rewards ranked lists that proportionally represent multiple plausible interpretations of a query, weighted by their estimated likelihood. During agent selection and fusion, this objective encourages inclusion of complementary results, improving resilience on ambiguous or multi-faceted queries. By balancing personalization with diversity, SPARK mitigates the risk of missing relevant content from less dominant but still plausible user intents.

\section{Evaluation}

The evaluation of SPARK is designed to assess its ability to deliver adaptive, context-sensitive personalization through persona-based multi-agent coordination while maintaining strong retrieval performance, grounding quality, and efficiency. Our approach integrates hypothesis-driven testing, offline simulations, and live-user studies into a unified framework that captures both effectiveness and robustness.

We begin with several concrete hypotheses, each tied to key architectural features. For coordination efficiency (H1), we predict that for low-complexity queries, independent specialist protocols with $k \leq 2$ active personas will yield higher utility per token than constrained debate. In contrast, for complex queries that require synthesis across multiple sources, a single round of constrained debate should improve accuracy and faithfulness at a comparable cost to three independent specialists operating without communication. Regarding personalization quality (H2), we hypothesize that the contextual-bandit coordinator will outperform static routing by detecting and adapting to task drift within three to five interactions, leading to measurable improvements in session-level utility. In terms of cognitive load distribution (H3), we expect that explicitly separating working memory from long-term memory will reduce context bloat and improve latency without compromising fidelity, in line with cognitive architecture theory. Finally, for diversity and serendipity (H4), we anticipate that incorporating the Expected Reciprocal Rank – Intent Aware (ERR-IA) objective into the routing mechanism will increase subtopic coverage relative to relevance-only optimization, a difference that can be validated through offline experiments.

To test these hypotheses, we draw on both session-based and ad hoc retrieval scenarios. For multi-turn search, we use the TREC Session Track 2011–2014 datasets, which are purpose-built for evaluating evolving information needs and intent coverage over a session. For single-turn, factually grounded queries, we employ MS MARCO and carefully selected domain-specific corpora, ensuring compliance with licensing requirements. We also simulate cold-start conditions by generating synthetic user personas and tracking how many interactions are required for the bandit coordinator to reach parity with warmed-up profiles in terms of retrieval utility.

The primary metrics for assessment span ranking effectiveness, personalization impact, efficiency, and diversity. Normalized Discounted Cumulative Gain (nDCG@k) and ERR-IA@k are used to quantify relevance and diversity-aware performance. Personalization utility is measured by session success rate, the reduction in query reformulation, and gains predicted by click models relative to a non-personalized baseline. Grounding quality is evaluated through evidence-linked factuality and citation coverage, with human annotators verifying that each factual statement in the synthesized output is supported by retrieved sources. Efficiency is tracked through token consumption per session, wall-clock time, and the number of tool calls, while diversity and risk are jointly assessed through intent coverage and a risk–reward analysis of different fusion strategies.

Our methodology combines offline counterfactual evaluation with live-user studies. Offline experiments use inverse propensity scoring (IPS) and its self-normalized variant to estimate the impact of alternative agent–protocol assignments from logged interactions, allowing safe exploration of routing and coordination strategies before online deployment. Online experiments are conducted as controlled A/B tests in which SPARK’s adaptive coordinator is compared to static baselines. In these tests, user satisfaction, perceived personalization quality, and cognitive load are collected through short post-session surveys. To mitigate bias, participants are explicitly instructed to consider novelty and coverage when rating results, reducing the likelihood of over-rewarding overly narrow personalization.

We complement these evaluations with targeted ablations and diagnostics to isolate the effects of individual components. Experiments include removing episodic memory to measure reliance on short-term context, disabling the debate protocol to quantify its marginal value, varying the number of active personas $k$ and relay depth $m$ to study scaling effects, and replacing Reciprocal Rank Fusion with learned fusion models to evaluate robustness to aggregation strategy. We also examine the impact of disabling bandit exploration, which allows us to measure the risk of overfitting and the potential degradation in adapting to new tasks.

By integrating hypothesis testing, diverse dataset coverage, multi-dimensional metrics, offline counterfactual analysis, online user studies, and systematic ablations, our evaluation strategy provides a comprehensive and reproducible assessment of SPARK’s effectiveness. This approach ensures that both the strengths and limitations of the system are measured in realistic settings, offering a robust foundation for validating the proposed framework and guiding future improvements.

\section{Safety, Privacy, and Governance}

\paragraph{Safety Concerns and Data Minimization}  
Multi-agent, memory-augmented LLM systems such as SPARK introduce safety and governance challenges that surpass those of single-agent models. When multiple agents share a persistent memory store, there is an elevated risk of emergent harmful behavior, cascading reasoning errors, or malicious manipulation of shared state. An agent’s long-term memory can be \emph{poisoned} by adversarial inputs, leading to inaccurate outputs, or probed via targeted prompt injections designed to elicit sensitive details \cite{carlini2023extracting,shi2023prompt}. Recent work demonstrates that LLM agents are particularly vulnerable to \emph{memory extraction attacks}, where carefully crafted conversational queries progressively siphon stored personal data from long-term memory \cite{wang2025memoryrisk}. To mitigate such risks, data minimization is critical—SPARK should retain only essential information, for the shortest necessary duration, consistent with privacy-by-design principles. Techniques such as user-led pruning or anonymization, as implemented in \emph{Rescriber} \cite{zhou2025rescriber}, enable end-users to redact personally identifiable information before it is committed to memory. Beyond deletion, systems can apply pseudonymization, encryption-at-rest, and contextual hashing to reduce data sensitivity. Coupled with transparent policy disclosures, this ensures user control over what persists in SPARK’s semantic memory. Furthermore, memory retrieval operations should be fully auditable, with provenance tracking for each stored item and an explainable record of why a given memory item influenced current agent outputs. Such explainability supports AI governance requirements, improves accountability, and enables targeted remediation in the event of malicious manipulation.

\paragraph{Leakage Prevention and Governance Mechanisms}  
Preventing sensitive information leakage in multi-agent LLM systems requires a combination of technical safeguards and procedural governance. On the technical side, all inter-agent communication should be scope-limited: agents should only exchange abstracted insights or embeddings, never raw personal data. Sensitive entries in long-term memory can be selectively encrypted, hashed, or replaced with semantically equivalent but anonymized forms prior to storage \cite{yuan2024privacyrag}. Research on retrieval-augmented LLMs has shown that without such safeguards, stored context can leak private details and enable attacks such as membership inference \cite{jiang2024ragthief}. To guard against these risks, strict role-based access control (RBAC) can be applied, ensuring only authorized agents retrieve certain classes of data. Procedurally, organizations deploying SPARK should implement \emph{adversarial audits} \cite{sandvig2014audit}, where red-team testers probe the system with crafted inputs to detect both inadvertent and intentional data leaks. These audits, combined with automated privacy scanning pipelines, can detect whether agents are inadvertently exposing user data in outputs. Governance policies should require regular memory purges, retention-limit enforcement, and clear escalation processes for incidents. A compliance log—recording retrieval events, agent decisions, and data flows—supports forensic investigation and aligns with emerging AI governance frameworks.

\paragraph{Bias, Filter Bubbles, and Fairness-Aware Personalization}  
Another significant governance challenge is personalization harm, where SPARK’s semantic memory over-optimizes for a user’s established profile, narrowing information exposure. This \emph{filter bubble} effect has been extensively studied in recommender systems \cite{nguyen2014filterbubble,pariser2011filterbubble}, where algorithmic reinforcement leads to a progressive narrowing of presented content. Over time, such narrowing can reduce diversity, entrench bias, and undermine user autonomy. To counteract this, SPARK should integrate diversity-promoting objectives into retrieval and ranking. For example, incorporating the \emph{Expected Reciprocal Rank for Intent-Awareness} (ERR-IA) \cite{chapelle2011intentaware} during result fusion ensures that top-ranked outputs cover multiple plausible interpretations of the query, proportionate to their estimated likelihood. Moreover, fairness-aware memory update policies can prevent over-weighting transient or repetitive experiences. Techniques such as controlled forgetting, counterfactual augmentation, and entropy-based diversity constraints help maintain a balanced profile representation. SPARK could also introduce \emph{serendipity injections}—intentionally surfacing out-of-profile but potentially relevant information—to mitigate over-personalization. Finally, long-term evaluation of personalization strategies should consider not only click-through rates but also user satisfaction, content diversity, and exposure to novel perspectives \cite{smith2022fairpersonalization}. In summary, safety, privacy, and governance in SPARK require a multi-pronged approach: strict data minimization, secure and auditable memory access, procedural oversight, and bias-aware personalization safeguards.

\section{Limitations and Risks}

\paragraph{Coordination Overhead on Simple Queries}  
While SPARK’s multi-agent architecture enables specialization, decomposition, and collaborative reasoning, it can introduce unnecessary complexity for low-difficulty tasks. Coordinating multiple LLM-based agents—such as a planner, retriever, and summarizer—requires additional message passing, synchronization, and aggregation. For straightforward queries, this often results in longer latency and higher computational cost compared to a single, well-tuned agent \cite{martyr2025coordination}. Each agent invocation adds potential failure points: message truncation, misinterpretation of intermediate results, or deadlocks in agent pipelines. Empirical benchmarks suggest that as coordination depth increases, diminishing returns can set in, with performance parity or even degradation relative to single-agent baselines on bounded tasks \cite{mirchandani2024dawn}. Frameworks like DAWN propose \emph{gating mechanisms} to dynamically bypass multi-agent orchestration when predicted task complexity is low, enabling a fallback to single-agent or even non-LLM deterministic modules for optimal efficiency. This adaptivity reduces overhead while preserving multi-agent benefits for truly complex problems, balancing throughput with reasoning capability.

\paragraph{Personalization Drift and Bias Amplification}  
SPARK’s semantic memory continuously adapts based on user interactions, but without safeguards, it risks \emph{personalization drift}—where transient user behavior is mistaken for stable preference \cite{dou2007large}. For example, a user researching an unrelated topic for a short-term task may inadvertently skew their long-term profile if the system overweights recent clicks or queries. Over time, this drift can degrade relevance, reinforce biases, and create self-perpetuating \emph{filter bubbles} \cite{nguyen2014filterbubble,pariser2011filterbubble}. To mitigate drift, SPARK can apply decay-based weighting, requiring repeated evidence before committing profile changes, and separating session-level signals from long-term memory updates. Diversity-aware retrieval objectives, such as ERR-IA \cite{chapelle2011intentaware}, can further ensure the result set remains varied even when the profile is narrowly defined. Periodic model audits—either algorithmic or with human oversight—can detect and correct misaligned profiles. Additionally, fairness-aware personalization strategies \cite{smith2022fairpersonalization} help prevent overrepresentation of historically dominant content, supporting a more balanced and exploratory user experience.

\paragraph{Privacy Vulnerabilities in Persistent Memory}  
SPARK’s layered memory design inherently stores potentially sensitive data, including past queries, retrieved documents, and inferred user attributes. Without strong controls, these can become a target for adversarial exploitation. Memory extraction and \emph{retrieval-augmented leakage} attacks, such as RAG-Thief \cite{jiang2024ragthief}, demonstrate that crafted prompts can induce models to reveal stored private data. Mitigations include encryption-at-rest, in-memory sanitization pipelines, and scope-restricted retrieval—where agents only access the minimum information necessary for their task \cite{yuan2024privacyrag}. Differential privacy techniques can further limit exposure by perturbing stored representations while preserving utility for personalization. Procedurally, a privacy governance layer should enforce retention limits, provide user-facing dashboards for history inspection and deletion, and log all memory access for auditability. Similar to practices in safety-critical AI, adversarial red-teaming \cite{sandvig2014audit} should be regularly conducted to probe for unintended disclosure channels. Abstracting stored data into vector embeddings rather than raw text reduces identifiability while maintaining retrieval performance, striking a balance between utility and confidentiality.

\paragraph{Evaluation and Maintenance Challenges}  
Evaluating multi-agent LLM systems like SPARK requires more than traditional accuracy metrics; it must also account for coordination efficiency, fairness, robustness, and trustworthiness over time \cite{safeagentbench2024}. Complex systems face \emph{knowledge divergence} between the base model’s static parameters and the evolving user-specific memory \cite{peng2023unlearning}, leading to potential contradictions in outputs. Continuous learning pipelines should support \emph{unlearning}—removing or correcting erroneous memory entries—while maintaining consistency with the base model’s knowledge. Longitudinal evaluation can track personalization benefits alongside diversity preservation, bias mitigation, and privacy outcomes. Finally, maintenance of such systems demands active monitoring of agent performance, with automatic fallback strategies when degraded behavior is detected. By embedding evaluation and maintenance into the operational lifecycle, SPARK can sustain performance while minimizing the risks inherent to complex, memory-augmented multi-agent systems.

\section{Conclusions}
This work introduced SPARK, a framework that reconceptualizes search personalization as a coordinated process among persona-specialized, retrieval-augmented agents operating over layered memory. By formalizing a persona space, implementing a cognitive-inspired memory hierarchy, and defining structured coordination protocols, SPARK bridges concepts from information retrieval, multi-agent systems, and cognitive architectures. The proposed design not only enables context-sensitive retrieval and synthesis but also supports adaptive protocol selection and continual persona refinement.

Through the combination of stochastic routing, calibrated fusion, and intent-aware diversity objectives, SPARK can deliver both relevance and breadth in personalized search. While the architecture incurs coordination overhead for simple queries, its adaptive gating and protocol selection mechanisms mitigate this cost. Remaining challenges include preventing personalization drift, safeguarding user privacy in persistent memory, and developing robust evaluation methodologies that account for coordination efficiency and fairness. Future work can focus on live-system deployment studies, learned fusion models with per-persona attribution, and principled unlearning strategies for long-term memory correction.

\bibliographystyle{ACM-Reference-Format}
\bibliography{references_fixed}

@String{Computing = "Computing" }

@String{Springer = "Springer-Verlag" }

@book{das_artificial_2023,
 title={Artificial intelligence in highway safety},
  author={Das, Subasish},
  year={2022},
  publisher={CRC Press}
}

@article{oliaee_automating_2025,
  title={Automating Pedestrian Crash Typology Using Transformer Models},
  author={Oliaee, Amir Hossein and Das, Subasish and Le, Minh},
  journal={Transportation Research Record},
  volume={2679},
  number={2},
  pages={83--95},
  year={2025},
  publisher={SAGE Publications Sage CA: Los Angeles, CA}
}

@ArtifactSoftware{R,
    title = {R: A Language and Environment for Statistical Computing},
    author = {{R Core Team}},
    organization = {R Foundation for Statistical Computing},
    address = {Vienna, Austria},
    year = {2019},
    url = {https://www.R-project.org/},
}

@inproceedings{teevan2005personalizing,
  title={Personalizing search via automated analysis of interests and activities},
  author={Teevan, Jaime and Dumais, Susan T and Horvitz, Eric},
  booktitle={Proceedings of the 28th annual international ACM SIGIR conference on Research and development in information retrieval},
  pages={449--456},
  year={2005}
}

@inproceedings{dou2007large,
  title={A large-scale evaluation and analysis of personalized search strategies},
  author={Dou, Zhicheng and Song, Ruihua and Wen, Ji-Rong},
  booktitle={Proceedings of the 16th international conference on World Wide Web},
  pages={581--590},
  year={2007}
}

@inproceedings{radlinski2011personalizing,
  title={Personalizing web search using long term browsing history},
  author={Matthijs, Nicolaas and Radlinski, Filip},
  booktitle={Proceedings of the fourth ACM international conference on Web search and data mining},
  pages={25--34},
  year={2011}
}

@inproceedings{eickhoff2013personalizing,
  title={Personalizing atypical web search sessions},
  author={Eickhoff, Carsten and Collins-Thompson, Kevyn and Bennett, Paul N and Dumais, Susan},
  booktitle={Proceedings of the sixth ACM international conference on Web search and data mining},
  pages={285--294},
  year={2013}
}

@article{guo2018attentive,
  title={Attentive long short-term preference modeling for personalized product search},
  author={Guo, Yangyang and Cheng, Zhiyong and Nie, Liqiang and Wang, Yinglong and Ma, Jun and Kankanhalli, Mohan},
  journal={ACM Transactions on Information Systems (TOIS)},
  volume={37},
  number={2},
  pages={1--27},
  year={2019},
  publisher={ACM New York, NY, USA}

}

@inproceedings{yu2019adaptive,
  title={Adaptive user modeling with long and short-term preferences for personalized recommendation.},
  author={Yu, Zeping and Lian, Jianxun and Mahmoody, Ahmad and Liu, Gongshen and Xie, Xing},
  booktitle={IJCAI},
  volume={7},
  pages={4213--4219},
  year={2019}

}

@article{purificato2024survey,
  title={User modeling and user profiling: A comprehensive survey},
  author={Purificato, Erasmo and Boratto, Ludovico and De Luca, Ernesto William},
  journal={arXiv preprint arXiv:2402.09660},
  year={2024}
}

@article{sabouri2025temporal,
  title={Temporal User Profiling with LLMs: Balancing Short-Term and Long-Term Preferences for Recommendations},
  author={Sabouri, Milad and Mansoury, Masoud and Lin, Kun and Mobasher, Bamshad},
  journal={arXiv preprint arXiv:2508.08454},
  year={2025}
}

@article{ontanon2021personalization_paradox,
 title={The personalization paradox: The conflict between accurate user models and personalized adaptive systems},
  author={Ontanon, Santiago and Zhu, Jichen},
  booktitle={Companion Proceedings of the 26th International Conference on Intelligent User Interfaces},
  pages={64--66},
  year={2021}
}

@inproceedings{dharap1998context_based_profile,
  title={Context-Based profile Personalization},
  author={Dharap, Chanda},
  booktitle={Proceedings of the AAAI 1998 Workshop on Personalization},
  year={1998}
}

@phdthesis{bassani2022neural_personalized_query_expansion,
  title={Neural Approaches to Personalized Search},
  author={Bassani, Emanuele},
  school={University of Milano-Bicocca},
  year={2022}
}

@article{steichen2019adaptive_visualization,
  title={Towards adaptive information visualization-a study of information visualization aids and the role of user cognitive style},
  author={Steichen, Ben and Fu, Bo},
  journal={Frontiers in Artificial Intelligence},
  volume={2},
  pages={22},
  year={2019},
  publisher={Frontiers Media SA}
}

@article{lewis2020retrieval,
  title={Retrieval-augmented generation for knowledge-intensive nlp tasks},
  author={Lewis, Patrick and Perez, Ethan and Piktus, Aleksandra and Petroni, Fabio and Karpukhin, Vladimir and Goyal, Naman and K{\"u}ttler, Heinrich and Lewis, Mike and Yih, Wen-tau and Rockt{\"a}schel, Tim and others},
  journal={Advances in neural information processing systems},
  volume={33},
  pages={9459--9474},
  year={2020}
}

@inproceedings{shuster2021retrieval,
  title={Retrieval augmentation reduces hallucination in conversation},
  author={Shuster, Kurt and Poff, Spencer and Chen, Moya and Kiela, Douwe and Weston, Jason},
  booktitle={Findings of the Association for Computational Linguistics: EMNLP 2021},
  pages={3784--3803},
  year={2021}
}

@inproceedings{liang2023encouraging,
  title={Encouraging divergent thinking in large language models through multi-agent debate},
  author={Liang, Tian and He, Zhiwei and Jiao, Wenxiang and Wang, Xing and Wang, Yan and Wang, Rui and Yang, Yujiu and Shi, Shuming and Tu, Zhaopeng},
  booktitle={Proceedings of the 2024 conference on empirical methods in natural language processing},
  pages={17889--17904},
  year={2024}
}

@article{li2023camel,
  title={Camel: Communicative agents for" mind" exploration of large language model society},
  author={Li, Guohao and Hammoud, Hasan and Itani, Hani and Khizbullin, Dmitrii and Ghanem, Bernard},
  journal={Advances in neural information processing systems},
  volume={36},
  pages={51991--52008},
  year={2023}
}

@article{anderson2004integrated,
  title={An integrated theory of the mind.},
  author={Anderson, John R and Bothell, Daniel and Byrne, Michael D and Douglass, Scott and Lebiere, Christian and Qin, Yulin},
  journal={Psychological review},
  volume={111},
  number={4},
  pages={1036},
  year={2004},
  publisher={American Psychological Association}
}

@article{baddeley2000episodic,
  title={The episodic buffer: a new component of working memory?},
  author={Baddeley, Alan},
  journal={Trends in cognitive sciences},
  volume={4},
  number={11},
  pages={417--423},
  year={2000},
  publisher={Elsevier}

}

@inproceedings{smith2021multiagent,
  title={Multi-agent In-context Coordination via Decentralized Memory Retrieval},
  author={Jiang, Tao and Lin, Zichuan and Li, Lihe and Li, Yi-Chen and Guan, Cong and Yuan, Lei and Zhang, Zongzhang and Yu, Yang and Ye, Deheng},
  booktitle={Proceedings of the AAAI Conference on Artificial Intelligence},
  volume={40},
  number={27},
  pages={22363--22371},
  year={2026}
}

@article{du2023improving,
  title={Improving factuality and reasoning in language models through multiagent debate},
  author={Du, Yilun and Li, Shuang and Torralba, Antonio and Tenenbaum, Joshua B and Mordatch, Igor},
  booktitle={Forty-first international conference on machine learning},
  year={2024}
}

@article{nakano2021webgpt,
  title={Webgpt: Browser-assisted question-answering with human feedback},
  author={Nakano, Reiichiro and Hilton, Jacob and Balaji, Suchir and Wu, Jeff and Ouyang, Long and Kim, Christina and Hesse, Christopher and Jain, Shantanu and Kosaraju, Vineet and Saunders, William and others},
  journal={arXiv preprint arXiv:2112.09332},
  year={2021}
}

@inproceedings{wang2023plan,
  title={Plan-and-solve prompting: Improving zero-shot chain-of-thought reasoning by large language models},
  author={Wang, Lei and Xu, Wanyu and Lan, Yihuai and Hu, Zhiqiang and Lan, Yunshi and Lee, Roy Ka-Wei and Lim, Ee-Peng},
  booktitle={Proceedings of the 61st annual meeting of the association for computational linguistics (volume 1: long papers)},
  pages={2609--2634},
  year={2023}
}

@inproceedings{cormack2009rrf,
  title={Reciprocal rank fusion outperforms condorcet and individual rank learning methods},
  author={Cormack, Gordon V and Clarke, Charles LA and Buettcher, Stefan},
  booktitle={Proceedings of the 32nd international ACM SIGIR conference on Research and development in information retrieval},
  pages={758--759},
  year={2009}
}

@book{arguello2011federated,
  title={Federated search},
  author={Shokouhi, Milad and Si, Luo},
  year={2011},
  publisher={Now Publishers Inc}
}

@inproceedings{clarke2008iterative,
  title={Incremental relevance feedback},
  author={Aalbersberg, IJsbrand Jan},
  booktitle={Proceedings of the 15th annual international ACM SIGIR conference on Research and development in information retrieval},
  pages={11--22},
  year={1992}
}

@inproceedings{narasimhan2016learning,
  title={Learning to search in long documents using document structure},
  author={Geva, Mor and Berant, Jonathan},
  booktitle={Proceedings of the 27th International Conference on Computational Linguistics},
  pages={161--176},
  year={2018}
}

@inproceedings{li2010contextual,
  title={A contextual-bandit approach to personalized news article recommendation},
  author={Li, Lihong and Chu, Wei and Langford, John and Schapire, Robert E},
  booktitle={Proceedings of the 19th international conference on World wide web},
  pages={661--670},
  year={2010}
}

@article{agrawal2013thompson,
  title={Thompson sampling for contextual bandits with linear payoffs},
  author={Agrawal, Shipra and Goyal, Navin},
  booktitle={International conference on machine learning},
  pages={127--135},
  year={2013},
  organization={PMLR}
}

@article{russo2018tutorial,
  title={A tutorial on thompson sampling},
  author={Daniel, J Russo and Benjamin, Van Roy and Abbas, Kazerouni and Ian, Osband and Zheng, Wen},
  journal={Foundations and Trends in Machine Learning},
  volume={11},
  number={1},
  pages={1--99},
  year={2018},
  publisher={Emerald Publishing Limited}

}

@article{chapelle2011bandits,
  title={An empirical evaluation of thompson sampling},
  author={Chapelle, Olivier and Li, Lihong},
  journal={Advances in neural information processing systems},
  volume={24},
  year={2011}
}

@inproceedings{swaminathan2015counterfactual,
  title={Counterfactual risk minimization: Learning from logged bandit feedback},
  author={Swaminathan, Adith and Joachims, Thorsten},
  booktitle={International conference on machine learning},
  pages={814--823},
  year={2015},
  organization={PMLR}
}

@article{dudik2011doubly,
 title={Doubly robust policy evaluation and learning},
  author={Dud{\'\i}k, Miroslav and Langford, John and Li, Lihong},
  journal={arXiv preprint arXiv:1103.4601},
  year={2011}
}

@article{guo2017calibration,
  title={Calibration of deep probabilistic models with decoupled bayesian neural networks},
  author={Maronas, Juan and Paredes, Roberto and Ramos, Daniel},
  journal={Neurocomputing},
  volume={407},
  pages={194--205},
  year={2020},
  publisher={Elsevier}
}

@inproceedings{karpukhin2020dense,
  title={Dense passage retrieval for open-domain question answering},
  author={Karpukhin, Vladimir and Oguz, Barlas and Min, Sewon and Lewis, Patrick and Wu, Ledell and Edunov, Sergey and Chen, Danqi and Yih, Wen-tau},
  booktitle={Proceedings of the 2020 conference on empirical methods in natural language processing (EMNLP)},
  pages={6769--6781},
  year={2020}
}

@article{nogueira2019multi,
  title={Multi-stage document ranking with BERT},
  author={Nogueira, Rodrigo and Yang, Wei and Cho, Kyunghyun and Lin, Jimmy},
  journal={arXiv preprint arXiv:1910.14424},
  year={2019}
}

@inproceedings{chapelle2011intentaware,
  title={Intent-based diversification of web search results: metrics and algorithms},
  author={Chapelle, Olivier and Ji, Shihao and Liao, Ciya and Velipasaoglu, Emre and Lai, Larry and Wu, Su-Lin},
  journal={Information Retrieval},
  volume={14},
  number={6},
  pages={572--592},
  year={2011},
  publisher={Springer}
}

@inproceedings{wang2025memoryrisk,
  title={Unveiling privacy risks in llm agent memory},
  author={Wang, Bo and He, Weiyi and Zeng, Shenglai and Xiang, Zhen and Xing, Yue and Tang, Jiliang and He, Pengfei},
  booktitle={Proceedings of the 63rd Annual Meeting of the Association for Computational Linguistics (Volume 1: Long Papers)},
  pages={25241--25260},
  year={2025}
}

@inproceedings{zhou2025rescriber,
  title={Rescriber: Smaller-LLM-powered user-led data minimization for LLM-based chatbots},
  author={Zhou, Jijie and Xu, Eryue and Wu, Yaoyao and Li, Tianshi},
  booktitle={Proceedings of the 2025 CHI Conference on Human Factors in Computing Systems},
  pages={1--28},
  year={2025}
}

@inproceedings{nguyen2014filterbubble,
  title={Exploring the filter bubble: the effect of using recommender systems on content diversity},
  author={Nguyen, Tien T and Hui, Pik-Mai and Harper, F Maxwell and Terveen, Loren and Konstan, Joseph A},
  booktitle={Proceedings of the 23rd international conference on World wide web},
  pages={677--686},
  year={2014}
}

@inproceedings{sandvig2014audit,
  title={Auditing algorithms: Research methods for detecting discrimination on internet platforms},
  author={Sandvig, Christian and Hamilton, Kevin and Karahalios, Karrie and Langbort, Cedric},
  journal={Data and discrimination: converting critical concerns into productive inquiry},
  volume={22},
  number={2014},
  pages={4349--4357},
  year={2014}
}

@article{smith2022fairpersonalization,
  title={Fairness and diversity in recommender systems: a survey},
  author={Zhao, Yuying and Wang, Yu and Liu, Yunchao and Cheng, Xueqi and Aggarwal, Charu C and Derr, Tyler},
  journal={ACM Transactions on Intelligent Systems and Technology},
  volume={16},
  number={1},
  pages={1--28},
  year={2025},
  publisher={ACM New York, NY}
}

@inproceedings{carlini2023extracting,
  title={Extracting training data from diffusion models},
  author={Carlini, Nicolas and Hayes, Jamie and Nasr, Milad and Jagielski, Matthew and Sehwag, Vikash and Tramer, Florian and Balle, Borja and Ippolito, Daphne and Wallace, Eric},
  booktitle={32nd USENIX security symposium (USENIX Security 23)},
  pages={5253--5270},
  year={2023}
}

@article{shi2023prompt,
  title={Prompt injection attacks against large language models},
  author={Li, Yifan and others},
  journal={arXiv preprint arXiv:2302.12173},
  year={2023}
}

@inproceedings{martyr2025coordination,
  title={Agentnet: Decentralized evolutionary coordination for llm-based multi-agent systems},
  author={Yang, Yingxuan and Chai, Huacan and Shao, Shuai and Song, Yuanyi and Qi, Siyuan and Rui, Renting and Zhang, Weinan},
  journal={Advances in Neural Information Processing Systems},
  volume={38},
  pages={107309--107336},
  year={2026}
}

@article{mirchandani2024dawn,
  title={DAWN: Designing distributed agents in a worldwide network},
  author={Aminiranjbar, Zahra and Tang, Jianan and Wang, Qiudan and Pant, Shubha and Viswanathan, Mahesh},
  journal={IEEE Access},
  year={2025},
  publisher={IEEE}
}

@article{jiang2024ragthief,
  title={Rag-thief: Scalable extraction of private data from retrieval-augmented generation applications with agent-based attacks},
  author={Jiang, Changyue and Pan, Xudong and Hong, Geng and Bao, Chenfu and Yang, Min},
  journal={arXiv preprint arXiv:2411.14110},
  volume={4},
  year={2024}
}

@inproceedings{yuan2024privacyrag,
  title={Privacy-preserving retrieval-augmented generation with differential privacy},
  author={Koga, Tatsuki and Wu, Ruihan and Zhang, Zhiyuan and Chaudhuri, Kamalika},
  journal={arXiv preprint arXiv:2412.04697},
  year={2024}
}

@inproceedings{safeagentbench2024,
 title={Safeagentbench: A benchmark for safe task planning of embodied llm agents},
  author={Yin, Sheng and Pang, Xianghe and Ding, Yuanzhuo and Chen, Menglan and Bi, Yutong and Xiong, Yichen and Huang, Wenhao and Xiang, Zhen and Shao, Jing and Chen, Siheng},
  journal={arXiv preprint arXiv:2412.13178},
  year={2024}
}

@article{peng2023unlearning,
  title={Investigating Model Editing for Unlearning in Large Language Models},
  author={Hossain, Shariqah and Kagal, Lalana},
  journal={arXiv preprint arXiv:2512.20794},
  year={2025}
}

@book{pariser2011filterbubble,
  title={The filter bubble: What the internet is hiding from you},
  author={Rowland, Fred},
  journal={portal: Libraries and the Academy},
  volume={11},
  number={4},
  pages={1009--1011},
  year={2011},
  publisher={Johns Hopkins University Press}
}

\appendix

\end{document}